\pdfoutput=1
\documentclass[11pt]{article}

\usepackage[final]{latex/acl}

\usepackage{times}
\usepackage{latexsym}

\usepackage[T1]{fontenc}
\usepackage[utf8]{inputenc}

\usepackage{microtype}

\usepackage{inconsolata}

\usepackage{graphicx}
\usepackage{subfigure}
\usepackage{listings}
\usepackage{makecell}

\usepackage{multirow}
\usepackage{diagbox} 
\usepackage[normalem]{ulem}
\usepackage{amsmath}
\usepackage{amsfonts}
\usepackage{amssymb}
\usepackage{mathtools}
\usepackage{booktabs}
\usepackage{xr}
\usepackage{hyperref}

\usepackage[normalem]{ulem}
\useunder{\uline}{\ul}{}
\usepackage[linesnumbered,ruled,vlined]{algorithm2e}
\SetKwInOut{Input}{Input}
\SetKwInOut{Output}{Output}
\SetKwInOut{Parameter}{Parameter}

\title{GenTKG: Generative Forecasting on Temporal Knowledge Graph\\with Large Language Models}

\author{Ruotong Liao$^{1, 2}$, \; Xu Jia$^{3}$, \; Yangzhe Li$^{3}$, \;  Yunpu Ma$^{1,2,4}$, \; \textbf{Volker Tresp}$^{1,2}$ \\
$^{1}$LMU Munich $\;$ $\;$  $^{2}$Munich Center for Machine Learning (MCML) $\;$ \\
$^{3}$Technical University of Munich $\;$ $^{4}$Siemens AG \\ 
\texttt{ruotong.liao@outlook.com, \; cognitive.yunpu@gmail.com}\\
\texttt{volker.tresp@lmu.de}
}

\begin{document}
\maketitle
% \addtocounter{footnote}{1}
% \footnotetext{Equal contribution.}
% \addtocounter{footnote}{1}
% \footnotetext{Work done prior to joining Amazon.}

\begin{abstract}

The rapid advancements in large language models (LLMs) have ignited interest in the temporal knowledge graph (tKG) domain, where conventional embedding-based and rule-based methods dominate. The question remains open of whether pre-trained LLMs can understand structured temporal relational data and replace them as the foundation model for temporal relational forecasting. Therefore, we bring temporal knowledge forecasting into the generative setting. However, challenges occur in the huge chasms between complex temporal graph data structure and sequential natural expressions LLMs can handle, and between the enormous data sizes of tKGs and heavy computation costs of finetuning LLMs. To address these challenges, we propose a novel retrieval-augmented generation framework named GenTKG combining a temporal logical rule-based retrieval strategy and few-shot parameter-efficient instruction tuning to solve the above challenges, respectively. Extensive experiments have shown that GenTKG outperforms conventional methods of temporal relational forecasting with low computation resources using extremely limited training data as few as 16 samples. GenTKG also highlights remarkable cross-domain generalizability with outperforming performance on unseen datasets without re-training, and in-domain generalizability regardless of time split in the same dataset. Our work reveals the huge potential of LLMs in the tKG domain and opens a new frontier for generative forecasting on tKGs. The code and data are released here:  \url{https://github.com/mayhugotong/GenTKG}.

\end{abstract}

\section{Introduction}
Forecasting the future lies in the intrinsic nature of humans to take controllability over the futural uncertainty ever since the existence of ancient fortunetellers who predict the future with insights into historical events. As the wave of Artificial General Intelligence (AGI) led by Large Language Models (LLMs)~\citep{bubeck2023sparks} showcases a persistent craving for 
% World Models\citep{matsuo2022deep} 
ability to model the complex information evolving in the real world, master the implicit rules and give predictions of what might happen next based on the historical observations~\citep{mialon2023augmented, matsuo2022deep}, we term this challenge for LLMs as \textit{Generative Forecasting}. We find Temporal Knowledge Graph (tKG) is a natural instance for investigating such a challenge attributed to the evolving world knowledge it contains and the task performed on it, namely \textit{temporal knowledge graph forecasting}. In short sentence, tKGs are multi-relational, directed graphs with labeled timestamped edges between entities (nodes) and can be viewed as streaming data sources where events come hourly, daily, or yearly, etc., and tKG forecasting task aims to forecast future events at timestamp $t$ based on past historical events before $t$. Specifically, tKG originates from Knowledge Graph (KG)~\citep{nickel2015review} which structures knowledge fact in the real world in the form of triples $(e_s, r, e_o)$, such as \textit{(Paris, the capital of, France)}, where $e_s, e_o$ represent the subject and object entity respectively, and $r$ represents the observed predicate between the two entities. As world knowledge evolves constantly over time such as the inaugurated presidents of the USA, the Temporal Knowledge Graph (tKG) was introduced by~\cite{tresp2015learning} to indicate the temporal effectiveness of the world events by extending a timestamp $t$ to form quadruples $(e_s, r, e_o, t)$. For example, \textit{(Donald Trump, the president of, the USA, 2021)} is followed by \textit{(Joe Biden, the president of, the USA, 2023)}.  The tKG forecasting task aims to answer queries $(e_s, r, ?, t)$ that predict the missing object given history events before $t$.%, which we visualize in Figure 1.

In tKG, the first embedding-based representation learning method is introduced by~\citep{ma2019embedding}. The following conventional embedding-based methods~\citep {goel2020diachronic, han2020explainable, sun2021timetraveler, yang2020re, li-etal-2022-hismatch} require carefully designed models that embed indexed quadruples into hidden latent space and hence lose the semantic aspects of events in tKGs. Besides, they require separate training for different datasets and hence suffer to handle even slight dataset modification and time split adaptation.
%Recent studies have hinted at the possibility of combining temporal knowledge graphs with pre-trained language models (PLMs) to enhance the temporal knowledge embeddings with language representations ~\citep{han2022enhanced}, however still costs complex constructions of carefully aligned structured graph data and unstructured textual data.
In stark contrast, the rule-based methods~\cite{liu2022tlogic} focus on mining temporal logic rules within the tKG graph structure in a symbolic way using neural networks. However, it possesses limited scalability to only similar datasets sharing similar rules. %Incorporating pre-trained language models(PLM) into TKGs has also shed light on the importance of the semantic aspects of TKGs.  utilize semantic aspects of TKGs, however, at the cost of constructing highly aligned graphs and textual data.  Which shed light on the importance of the semantic aspects of TKGs.
%These fragmented
With the huge advancements emerging with numerous large language models (LLMs)~\citep{wei2022emergent}, for example utilizing the emergent in-context learning (ICL) ability of LLMs~\citep{dong2022survey} by sequentializing temporal ascending ordered tKG facts to texts but failed to compete with the above conventional methods~\citep{lee2023temporal}. %Otherwise, bridging LLMs with TKGs remains under-explored.
The question remains open:  \textbf{Can pre-trained LLMs understand structured temporal relational data and replace conventional methods as the foundation model for temporal relational forecasting?}

To address the above issue, we bring temporal knowledge forecasting into the \textbf{\textit{generative forecasting} }setting and deliberately prioritize the most influential factors in these two domains: the temporal and structural characteristics of tKGs and the flexible natural language processing abilities of Large Language Models (LLMs). However, two challenges stand in the middle how to integrate them organically. The first is the \textbf{\textit{modality challenge}} between data structures. As tKG are complex temporal multi-relational graph data with tens of thousands of quadruples, it is hard to adapt to sequential natural language expressions that LLMs can process. The second is the \textbf{\textit{computation challenge}} with the enormous costs of fine-tuning LLMs especially with tens of thousands of quadruples requiring months of training time on consumable graphic cards. %However, open-source and consumable training remains a core demand for advancing forefront research.

To solve the above two challenges, we propose \textbf{GenTKG}, a novel \textit{retrieval-augmented generation} framework that solves the tKG forecasting task in the \textit{generative forecasting} setting, outperforming embedding-based, rule-based and ICL methods. Besides, GenTKG serves as an instantiation that sheds light on the promising \textit{generative forecasting} ability of LLMs. For the first \textit{modality challenge} between structured temporal graph data and sequential natural languages, we solve it in the retrieval phase. We utilize a temporal logical rule-based retrieval strategy (TLR) that mines the temporal logic rules of the tKGs and forms a rule bank. These rules serve to retrieve the most temporally and logically relevant historical facts to the given query. These facts are then sequentialized to natural languages in the ascending temporal order and filled in a specialized prompt template for LLMs. Although the prompts are in the form of sequential natural languages, they inherit structural information in the tKG implicitly since the extraction process is highly dependent on learned structural rules. 
These prompts enable LLMs to comprehend temporal relational data, and TLR enables the input window of LLM to serve as the implicit and decouplable interface for communicating temporal and structural relational data to LLMs. Moreover, TLR delivers improvement over the recent pure ICL method, regardless of the backbone LLM being used. %by introducing the retrieval phase on different LLMs only. 

For the second \textit{computation challenge} between huge tKG size and high computation costs of LLM, we solve it in the generation phase. We propose a few-shot parameter-efficient instruction-tuning strategy (FIT) that aligns LLM with a temporal relational forecasting task and reforms it into an autoregressive generation task.  We further decompose the second \textit{computation challenge} in two subtasks from the perspective of model and data respectively. The first subtask is to deal with the enormous computation costs and hardware requirements in training LLM. We solve this subtask with a parameter-efficient fine-tuning (PEFT) adaptation method, specifically Low-rank Adaptation (LoRA)~\citep{hu2021lora}. The second subtask is to deal with the enormous size of training data in tKGs. We deliberately think out of the box by bypassing learning the data like conventional methods and instead, letting the LLM learn the generative forecasting task on tKG. In other words, we reform data-centric model learning to task-centric LLM alignment that aligns LLMs with tKG forecasting task through instruction tuning. We have specially designed task instructions, retrieved facts as input, and generative predictions as output. Besides, we introduce few-shot tuning that further reduces training data to only 1024 prompt-response pairs which is as few as 0.27\%  of original tens of thousands of training data with exceeding performance. Under extreme cases, we could further reduce to as few as 16 samples which is 0.0042\% of original data while maintaining comparable performance to conventional methods.

% Specifically, we introduce efficient instruction-tuning technology with only 1024 carefully curated prompts and responses on a single consumable GPU. Our approach demonstrates remarkably strong performance over conventional rule-based and embedding-based models and reaches comparable results with even as few as 16 training samples. Moreover, our novel paradigm showcases a remarkable generalization ability that tends to generalize well to unseen TKG datasets without retraining like conventional methods.

Our approach offers a foundational framework for future explorations in generative forecasting on temporal knowledge graphs. Our contributions are:
% can be summarized as follows:
\begin{enumerate}
    \item \textbf{Opening a frontier of generative forecasting on tKG.} To the best of our knowledge, we are the first to introduce instruction-tuned generative LLM to the tKG domain. Our framework \textbf{GenTKG} proposes a novel retrieval augmented generation paradigm for tKG forecasting, regardless of the backbone LLM.
    \item \textbf{Drastically low computation costs with exceeding performance.} With only 16-shots parameter-efficient instruction tuning, we can already reach comparable results to conventional methods. With 1024-shots tuning, we can outperform existing rule-based, embedding-based, and the recent in-context-learning method.
    \item \textbf{Task reformulation from data learning to task alignment.}  We bypass designing specific models to learn specific tKG datasets. Instead, we novelly reform the data-centric learning to task-centric LLM alignment that aligns LLMs to generative forecasting task on tKG.
    \item \textbf{Generalizability across datasets without retraining}. With one-time training on a single dataset, our GenTKG has showcased remarkably both cross-domain and in-domain generalizability with exceeding performance on multiple datasets without retraining. 
\end{enumerate}

\section{Generative Forecasting on Temporal Knowledge Graph}
% \todo[an example of TLR]
% In this section, we describe our GenTKG framework following the two-phase Retrieval-then-Generate in separate sections. In section \ref{sec: retrieval}, we describe our retrieving strategy which firstly generates the temporal-logic rule-based retrieval strategy for retrieving highly temporally relevant and logic-following history facts. In section section \ref{}, we describe the few-shot parameter-efficient instruction-finetuning that aligns LLMs to generative forecasting on temporal knowledge graphs.

\begin{figure*}
    \centering
    \includegraphics[width=0.9\linewidth]{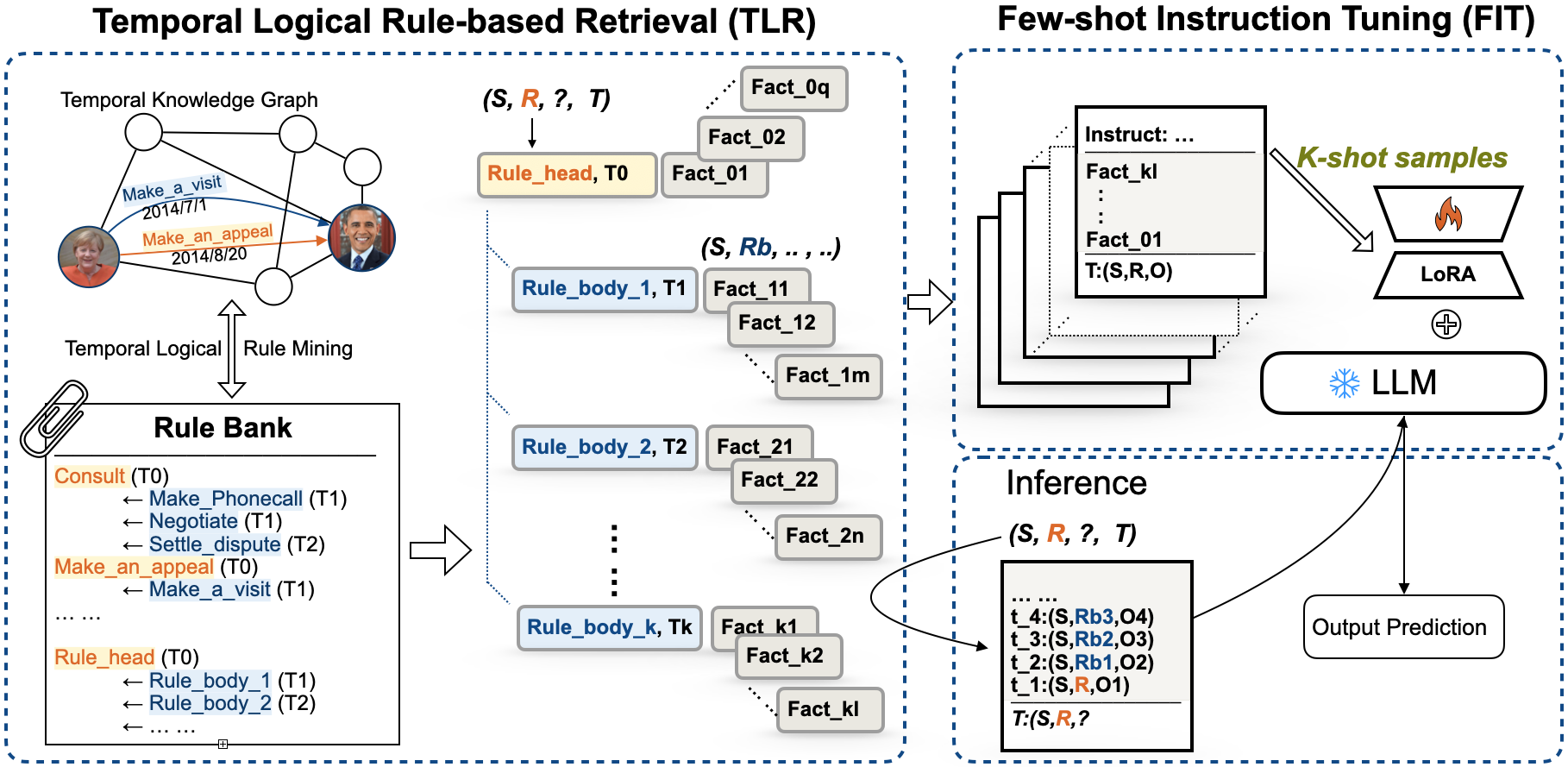}
    \caption{Framework of GenTKG. GenTKG first retrieves relevant facts based on a temporal logical rule-based retrieval strategy (TLR) then samples $K$ prompts for few-shot parameter-efficient instruction-tuning (FIT) that aligns LLM to the task of generative temporal knowledge graph forecasting.}
    \label{fig:main}
\end{figure*}

In this section, we explain our GenTKG framework following its two-phase methodology: Retrieve-then-Generate, in two sections.
%TODO: describe the reason
In Section \ref{sec:retrieval}, we explain the retrieval phase,  which proposes a temporal logical rule-based retrieval strategy (TLR) to capture historical facts that exhibit high temporal relevance and logical coherence. In Section \ref{sec: finetuning}, we delve into the details of the few-shot parameter-efficient instruction-finetuning strategy (FIT), an essential component that aligns Large Language Models (LLMs) to the task of generative forecasting on temporal knowledge graphs.

\subsection{Temporal Logic Rule-based Retrieval}\label{sec:retrieval} % can I mention first-order logic?
The TLR retrieval strategy is inspired by the phenomenon that a pair of entities can have many interactions at different timestamps such as a president visiting the same country multiple times. Another intuition behind this is that some relations tend to be temporally and logically sequential, for example in ICEWS14 we can see \textit{(Angela Merkel, discuss by telephone, Barack Obama, 2014/07/22)} and \textit{(Angela Merkel, consult, Barack Obama, 2014/08/09)}. Therefore, we borrow a partial idea of TLogic~\citep{liu2022tlogic} that mines the temporal logic rules hidden in the tKG structure. Notably, we opt to choose first-order temporal logic that complies with the input context constraints of the LLMs, and don't apply rules directly for ranking each entity as it did. Then we propose the novel TLR that retrieves the most temporally related and logically supportive history events for the given query based on these learned rules. To help understand our retrieval strategy, two definitions and the algorithm are given in the following. 

\paragraph{\textit{Definition I (Temporal Random Walk)}}
% \textit{\textbf{Definition I (Temporal Random Walk)}}
A non-increasing temporal random walk $W$ starting from subject entity $e_{s}\in \mathcal{E}$ to object entity $e_{o}\in \mathcal{E}$ in the tKG $\mathcal{G}$ is defined as a cycle of edges $((e_s, r_1, e_o, t_2),(e_s, r_2, e_o, t_1))$
with $t_2 > t_1$ where $(e_s, r_i, e_o, t_i) \in \mathcal{G}$ and $i \in {1, 2}$. The time constraints ensure that the edges are traversed only backward in time.%, where it is also possible to walk along edges with the same timestamp.

\paragraph{\textit{Definition II (Temporal Logical Rule)}} 
A cyclic temporal logical rule $R$ is defined
as $(E_1, r_h, E_2, T_2) \leftarrow (E_1, r_b, E_2, T_1)$ with $T_2 > T_1$, where $E_i$ and $T_i$ for $i \in {1,2}$ are replaceable variables that represent entities and timestamps. The left-hand side of $R$ is called the rule head, with $r_h$ being the head relation, while the right-hand side is called the rule body, with $r_b$ being the body relation. A rule head can be supported by multiple rule bodies denoting different rules as $\mathcal{TR}$. A $\mathcal{TR}$ implies that if the rule body holds then the rule head is true for
a future timestamp $T_2$. %The replacement of the variables $E_i$ and $T_i$ by constant terms is called grounding or instantiation. For example, a grounding of the temporal rule $(E1, consult, E2, T2) \leftarrow (E1, discuss\ by\ telephone, E2, T1)$ is given by the edges.  To estimate the probability of a rule's correctness,
The confidence of a rule conf$(\mathcal{TR})$ is defined as dividing the rule support by the body support, where the support is the number of quadruples satisfying rule bodies or rule heads with time constraints within $\mathcal{TR}$ . % with constraints $t_1 \leq t_2$. % $(e_1, r_b, e_2, t_1)$ satisfying the rule bodies and the rule support is the number of quadruples $(e_1, r_h, e_2, t_2)$ satisfying the rule head 

\paragraph{Rule Learning}
% Firstly, cyclic temporal walks are extracted from the TKG $\mathcal{G}$.
Let $r_h$ be a fixed relation, for which we want to learn rules. We sample an edge $(e_1, r_h, e_2, t)$, which will serve as the rule head, uniformly from all edges with relation $r_h$. Then the temporal random walker samples iteratively candidate edges adjacent to the current object $\mathcal{C}(e_2, t) \coloneqq \left\{\left(e_2, r, e_1, \hat{t}\right)\mid\left(e_2, r, e_1, \hat{t}\right) \in \mathcal{G}, \hat{t}<t\right\}$, where $\hat{t}$ is the timestamp associated with the next transition edge. Besides, we use an exponentially weighted transition distribution that prioritizes temporally closer edges during sampling which is defined as
\begin{equation}
\mathbb{P}\left(u ; e_2, t\right)=\frac{\exp \left(t_u-t\right)}{\sum_{\hat{u} \in \mathcal{C}\left(e_2, t\right)} \exp \left(t_{\hat{u}}-t\right)}
\end{equation}
where $t_u$ denotes the timestamp of edge $u$.
%A full temporal walk can be collected as $((e_1, r_h, e_2, t_1),(e_1, r_b, e_2, t_2))$. 
After a fixed sampling we can collect a set of temporal walks satisfying the rule $(E_1, r_h, E_2, T_2) \leftarrow (E_1, r_b, E_2, T_1)$.
We then estimate the confidence of the rules following the definition II. 
%Specifically, we sample a fixed number of body groundings that match the body relations and temporal constraints. 
%Specifically, we count the number of body supports that there exists a relation head $r_h$ connects $e_1$ and $e_2$ with the latest timestamp than all body timestamps.

\paragraph{Temporal Logic Rule-based Retrieval}
After gaining learned temporal logical rule sets, we order them according to the associated confidence scores. For a given forecast query $(e_s, r, ?, t)$ we retrieve a candidate subgraph $\mathcal{G}_s(e_s, r, t)$ from the TKG $\mathcal{G}$ containing temporally and logically relevant histories for the given query, with respect to the subject entity, relation, and timestamp. Since the query subject entity is fixed, there are two key factors in the retrieval algorithm, i.e. time window and rule grounding. First, we define the time window as $TW=[t_{-},t]$ with $t_{-} \coloneqq t-w$, where the $w \in \mathbb{N}^+$ represents the time window length backward starting from the query timestamp. The maximum length of $w$ is $\min{\{t_{max},t\}}$ with $t_{max}$ denoting the maximum timestamp of the datasets. Second, the query relation  $r$ is fixed as a rule head  $r_h$. Within each $TW$, we first use rule-head to retrieve history facts satisfying $(e_s, r_{h}, e_o, t-w)$. Then, we apply the learned rules $\mathcal{TR}$ and select top $k$ various rule bodies ${r_{b_1}, r_{b_2}, \cdots}, r_{b_k}$ regarding $r$ in descending confidence and add historical events $(e_s, r_{b}, e_o, t-w)$ to $\mathcal{G}_s(e_s, r, t)$ for the given query. The size of $\mathcal{G}_s(e_s, r, t)$ can be adjusted dynamically with respect to $w$ and $k$. We stop the retrieval until a maximum history length $N$ is reached. For instance, we retrieve history events iteratively with the descending confident rule bodies for each time window backtrace step until a maximum history length of 50 is reached.  At the end of the retrieval phase, we reorder all history events in temporal descending order for each query. The pseudo-code is attached in Appendix \ref{sss: ap_alg}.

% \subsection{Prompt Construction for In-Context Learning}

\subsection{Align LLM to Generative tKG Forecasting}\label{sec: finetuning}
The second phase of the proposed GenTKG framework contributes to transforming the conventional data-centric tKG model learning task into an alignment task that aligns LLM with generative forecasting on tKGs. % This study formulates the task as efficient training of a generative temporal forecasting model with a limited length of historical events from the last retrieval phase as the input. 
We utilize a few-shot parameter-efficient instruction tuning strategy (FIT) under the settings of low GPU resource consumption with a single graphic card. In \ref{sss:IPD}, we describe the instruction prompt design. In \ref{sss:PEFT}, we describe the parameter-efficient instruction tuning for training our generative model. In \ref{sss:FST} , we explain the few-shot tuning strategy that efficiently performs alignment with as few as 1024 samples and explores the lower-bound of samples for few-shot tuning. In \ref{sec:ind}, we describe the generalization ability of generative forecasting on tKG.

\subsubsection{Instruction Prompt Design}\label{sss:IPD}

\begin{figure}
    \centering
    \includegraphics[width=1.0\linewidth]{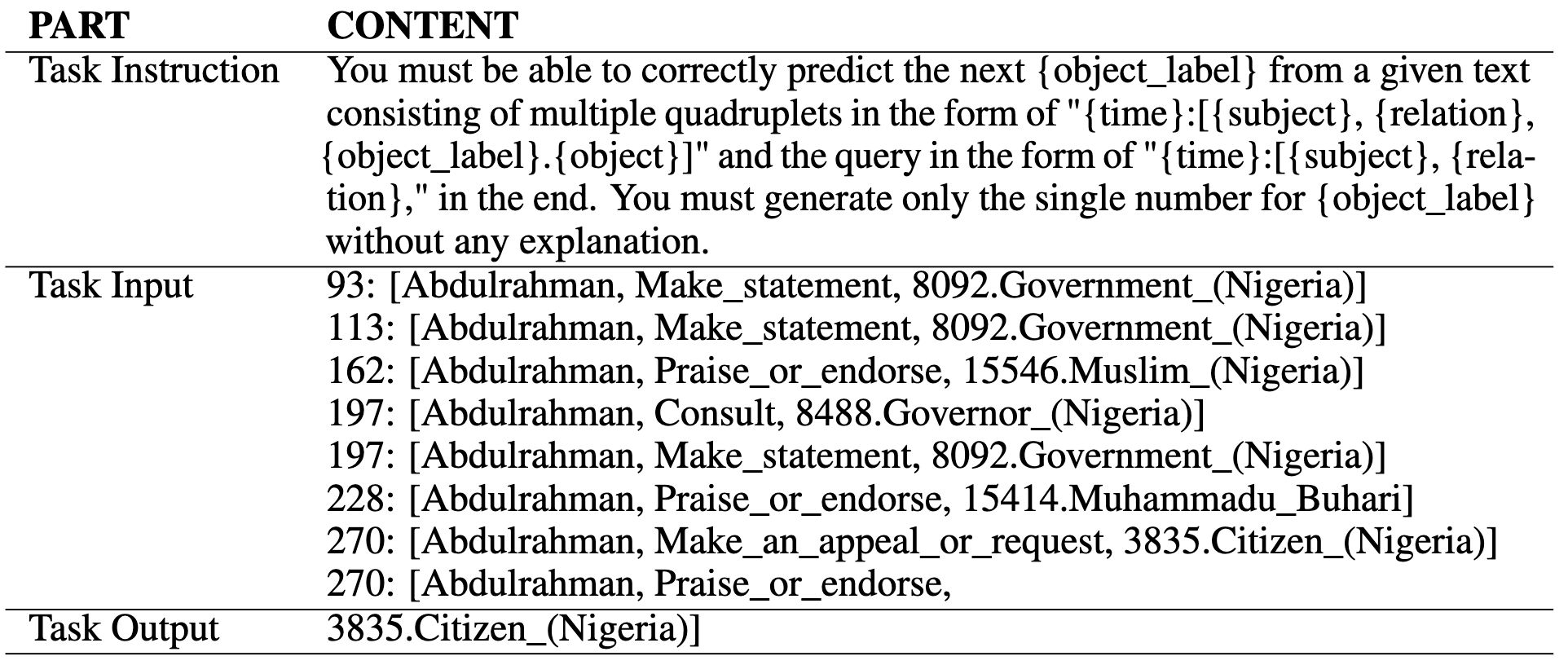}
    \caption{Instruction Prompt Design}
    \label{fig:IPD}
\end{figure}

Instruction Tuning is a crucial technique that fine-tunes LLMs with human-curated instruction and response pairs as the training data, empowering LLMs with instruction-following capability~\cite {zhou2023lima}.
The construction of an instruction sample is usually composed of three parts, i.e. task instruction, task input, and task output. Task instruction clarifies the definition of the task for LLMs to comprehend and gives explicit solutions for LLMs to follow and execute. Task input in natural languages is input data serving as context for LLMs.  Task output is the decoding results based on the input prompt.
In our proposed GenTKG framework, we adapt the temporal knowledge graph forecasting task to the instruction task for LLMs with individual adaptation for the three parts partially following the setting in \cite{lee2023temporal}. The instruction is depicted in Figure \ref{fig:IPD}. Except for the designed task instruction, the task input is modeled as ordered historical events retrieved from the TLR phase for a given query $(e_s, r, e_o, t)$ as described in \ref{sec:retrieval}. Each fact is filled in the template of ``$t:[e_s, r, n_{e_o}.e_o]$``. The query $(e_s, r, e_o, t)$ is expressed in a similar but partial way as ``$t:[e_s, r, $`` for LLM to complete as generative predictions. It is worth noting that we conserve the format in ~\citep{lee2023temporal} that maps each candidate object $e_o$ with a numerical index $ n_{e_o}$ as a fair comparison. However, \citep{lee2023temporal} try to avoid unfair tokenization for different entities with this index and use the probabilities of index tokens generated by the LLMs to get ranked scores of output entities in an indirect way. But this can only be used on GPT-like model and cannot handle LLaMA-like models harnessing individual tokenization.  Therefore we use top generated entity names directly for prediction evaluation.

\subsubsection{ Parameter-efficient Instruction Tuning} \label{sss:PEFT}
Direct fine-tuning of the entire model is computationally demanding and time-consuming. To address these computational challenges, we adopt 
%a lightweight fine-tuning strategy to align the LLM with tKG forecasting tasks.
%As language models often possess an excessive number of parameters in billion-sized, with the bulk of their information concentrated in a low intrinsic dimension, prior research has managed to achieve comparable performance to that of the full model by fine-tuning only a tiny additional subset of parameters. 
%Specifically, we implement 
the Low-Rank Adaptation (LoRA) technique~\citep{hu2021lora}. LoRA involves the freezing of pre-trained model parameters $\boldsymbol{\theta_0}$ while introducing trainable additional parameters $\boldsymbol{\theta_0}$ that can be decomposed into low-rank matrices $\Delta \boldsymbol{\theta}_0=\mathbf{B} \mathbf{A}, \mathbf{B} \in \mathbb{R}^{d \times r}, \mathbf{A} \in \mathbb{R}^{r \times k}, r\ll \min(d,k)$ that incorporat supplimentary information to the LLM.
%During adaptation, $\boldsymbol{\theta}_0$ is frozen while $\mathbf{B}$ and $\mathbf{A}$ are updated. By optimizing these rank decomposition matrices, supplementary information is incorperated efficiently into the frozen model. %while keeping the original parameters in a fixed state. %This approach enables us to strike a balance between computational efficiency and model performance, making aligning LLMs with temporal relational forecasting possible.

At present, there are large amounts of LLMs released, such as GPT series~\citep{kojima2022large, radford2019language}, T5 series~\citep{raffel2020exploring}, CHinchilla~\citep{hoffmann2022training}, and
LLaMA~\citep{touvron2023llama}, etc.. Among these, proprietary models can only be accessed by APIs such as ChatGPT with limited adaptation and alignment possibilities that hinder the research purpose. To facilitate the research of generative forecasting on temporal knowledge graph, we carefully opt for the open-sourcing LLMs, i.e. GPT-NeoX-20B~\citep{gpt-neox-20b} and LLaMA2-7B~\cite{touvron2023llama}, which is the third-party reproduction of GPT-3 and open-source public model respectively. Due to hardware limitations, we leave GPT-NeoX-20B frozen to investigate the effectiveness of our retrieval phase through its in-context learning ability. We perform the whole GenTKG framework on LLaMA2-7B with consumable adaptation. 

\subsubsection{Efficient Alignment with Few-shot Tuning } \label{sss:FST}
Our framework contributes a remarkably efficient and effective few-shot training strategy. The hypothesis has been proven that alignment can be a simple process where the LLMs learn the style or format for responding to prompts and expose the knowledge and capabilities that were already acquired during pretraining~\cite{zhou2023lima}. 
% Besides, limited human-curated prompts-response pairs can achieve comparable performances without large-scale instruction tuning in instruction following and dialogue scenarios. 
Therefore, considering the volume of temporal knowledge graphs that usually possess tens of thousands of training data, we propose a $K$-shot tuning paradigm where only an extremely limited number of $K$ samples are uniformly sampled from the temporal-ordered training set for language model adaptations. In our case, we select only 1024 samples which takes up as few as 0.27\% of the original GDELT dataset sizes that conventional methods usually fully trained on. We further prove that our method can acquire temporal relational forecasting capability rapidly with severely limited training data (0.0027\%) with an extreme 16-shot training setting while maintaining comparable performances to conventional methods.

\subsubsection{Generalization Ability of GenTKG} \label{sec:ind}
Due to the novel transformation from data-centric learning to task-centric alignment which forces the LLM is aligned to the temporal relational forecasting task itself rather than the learning of the tKG data. GenTKG also delivers remarkable generalizability in various generalization settings.

\textbf{Cross-domain generalizability.}
LLM trained on one dataset can be inferred directly on other datasets. A generalized GenTKG only requires learning the temporal-logical rule-based retrieval strategy for the new datasets in the first phase to ensure proper prompts with relevant histories. However, it doesn't require retraining LLM in the second phase. Still, high-performance gains are maintained and even comparable to the original setting. 

\textbf{In-domain generalizability}.
GenTKG maintains high-performance gains on the same dataset even trained on only partial training data. The partition can be limited to a small fraction such as 5\% of original training data. This characteristic exceeds conventional methods which always suffer drastic performance drops even with a minor change of critical value of the forecasting timestamp between the train and evaluation set. %We attribute this ability to the reason that the LLM in is not fine-tuned specifically to a single dataset. Instead, the LLM is aligned to the temporal relational forecasting task itself rather than the learning of the tKG data.

\section{Experimental Setup}
In this section, we describe the experimental setup of GenTKG framework. Specifically, we describe four datasets, the evaluation protocols, and the experimental design.

\textbf{Datasets}
Four benchmark datasets are used to evaluate
GenTKG: 1) ICEWS14~\citep{boschee2015icews}  2)
ICEWS18~\citep{boschee2015icews} 3) GDELT ~\citep{leetaru2013gdelt} 4) YAGO ~\citep{mahdisoltani2013yago3}. The two versions of the Integrated Crisis Early Warning System (ICEWS) both consist of timestamped political events, e.g., (Angela Merkel,
Visit, India, 2015-03-25). %The difference between the two datasets is the timespan, where ICEWS14 contains events occurring in 2014, while ICEWS18 contains events from January to October in 2018. The events in each dataset are granularized into days. 
The GDELT and YAGO datasets are extracted from the subsets of GDELT and YAGO knowledge bases containing facts and time information. Dataset statistics is shown in Table \ref{tab: dataset} in the Appendix \ref{sss: ap_dataset stat}.
%WIKI and YAGO are two subsets extracted from
%Wikipedia and YAGO3 (Mahdisoltani et al., 2013),
%respectively. The details of each dataset and the
%dataset split strategy are provided in Table\ref{}.\\

\textbf{Evaluation}
    % - Metric: Hits@1, Hits@3, Hits@10 \\
Since GenTKG generates entity predictions directly, we use the temporal-aware filtered~\cite{gastinger2023comparing} Hits@1/3/10 metric to evaluate extrapolated link prediction. Hits@1/3/10 denotes the proportion of the actual missing entities ranked within the top 1/3/10. 
% We report results based on the more strict implementations of raw metrics, where the entities that are valid predictions other than the exact query ground truth will not be removed from the ranking.

\textbf{Baselines}
Since GenTKG is the first method to introduce instruction-tuned generative models into the tKG forecasting domain, it is necessary to include three typical types of existing methods as baselines. The first are embedding-based methods, represented by RE-GCN~\citep{li2021temporal}, xERTE~\citep{han2020explainable}, TANGO~\citep{han2021learning}, and Timetraveler~\citep{sun2021timetraveler}. The rule-based method is TLogic~\cite{liu2022tlogic} and the third type is the LLM-based ICL method with frozen parameters~\cite{lee2023temporal}. %We provided detailed descriptions for each method in the Appendix\ref{}.

\textbf{Experiment Design}
In order to comprehensively analyze GenTKG compared to different conventional methods, there are three research questions to be answered. \textbf{RQ1}: How is the overall performance of the proposed GenTKG framework compared with the existing conventional embedding-based, rule-based TKG methods and LLM-based ICL method? \textbf{RQ2}: How well is the cross-domain and in-domain generalizability of GenTKG on different settings? \textbf{RQ3}: How do the components of the GenTKG affect its effectiveness?

\section{Experimental Results}
\externaldocument{10_appendix}

\begin{table*}[htbp]
\caption{Temporal link prediction results on temporal-aware filtered Hits@1/3/10(\%). The best results among each metric are highlighted in \textbf{bold} and the second bests are \underline{underlined}. %The standard errors of the fusion models are also provided.
}\label{tab: link prediction results PART1}
% YZ: 改过
\begin{center}
\resizebox{1.0\textwidth}{!}{
% Please add the following required packages to your document preamble:
% \usepackage{multirow}
% \usepackage{diagbox} 
% \usepackage[normalem]{ulem}
% \useunder{\uline}{\ul}{}\diagbox{Eval}{Train}
\begin{tabular}{l|l|ccc|ccc|ccc|ccc}
\toprule
\multirow{2}{*}{Method Type} & \multirow{2}{*}{\diagbox{Models}{Datasets}} & \multicolumn{3}{c|}{ICEWS14} & \multicolumn{3}{c|}{ICEWS18} & \multicolumn{3}{c|}{GDELT} & \multicolumn{3}{c}{YAGO} \\ \cline{3-14} 
 &  & Hits@1 & Hits@3 & Hits@10 & Hits@1 & Hits@3 & Hits@10 & Hits@1 & Hits@3 & Hits@10 & Hits@1 & Hits@3 & Hits@10 \\ \midrule
\multirow{4}{*}{Embedding-based} & RE-GCN & 31.3 & 47.3 & \textbf{62.6} & 22.3 & {\ul 36.7} & \textbf{52.5} & 8.4 & 17.1 & 29.9 & 46.8 & 60.7 & 72.9 \\
 & xERTE & 33.0 & 45.4 & 57.0 & 20.9 & 33.5 & 46.2 & 8.5 & 15.9 & 26.5 & 56.1 & 72.6 & 78.9 \\
 & TANGO & 27.2 & 40.8 & 55.0 & 19.1 & 31.8 & 46.2 & 9.4 & 18.9 & 32.2 & 56.6 & 65.1 & 71.8 \\
 & Timetraveler & 31.9 & 45.4 & 57.5 & 21.2 & 32.5 & 43.9 & 11.2 & 18.6 & 28.5 & 60.4 & 77.0 & 83.1 \\ \midrule
Rule-based & TLogic & 33.2 & {\ul 47.6} & {\ul 60.2} & 20.4 & 33.6 & {\ul 48.0} & 11.3 & {\ul 21.2} & \textbf{35.1} & 63.8 & 65.0 & 66.0 \\ \midrule
\multirow{2}{*}{ICL} & GPT-NeoX-20B & 32.6 & 44.0 & 54.2 & 18.2 & 29.5 & 41.4 & 6.8 & 12.0 & 21.1 & 72.6 & {\ul 81.0} & {\ul 84.6} \\
 & Llama2-7B & 25.8 & 43.0 & 51.0 & 13.5 & 27.6 & 32.6 & 3.6 & 12.5 & 22.0 & 67.7 & 79.0 & 81.8 \\ \midrule
\multirow{3}{*}{GenTKG} & GPT-NeoX-20B + TLR & {\ul 35.0} & 47.4 & 57.5 & 21.1 & 33.9 & 45.6 & 10.2 & 16.7 & 27.3 & {\ul 73.6} & \textbf{83.0} & \textbf{86.8} \\
 & Llama2-7B + GenTKG & \textbf{\begin{tabular}[c]{@{}c@{}}36.85 ±\\ 0.75\end{tabular}} & \textbf{\begin{tabular}[c]{@{}c@{}}47.95 ±\\ 0.75\end{tabular}} & \begin{tabular}[c]{@{}c@{}}53.5 ±\\ 0.8\end{tabular} & \textbf{\begin{tabular}[c]{@{}c@{}}24.25 ±\\ 0.75\end{tabular}} & \begin{tabular}[c]{@{}c@{}}{\textbf{37.25}} ±\\ {\textbf{0.25}}\end{tabular} & \begin{tabular}[c]{@{}c@{}}42.1 ±\\ 1.1\end{tabular} & \textbf{\begin{tabular}[c]{@{}c@{}}13.9 ±\\ 0.5\end{tabular}} & \textbf{\begin{tabular}[c]{@{}c@{}}22.55 ±\\ 0.55\end{tabular}} & \begin{tabular}[c]{@{}c@{}}{\ul 30.45} ±\\ {\ul 0.45}\end{tabular} & \textbf{\begin{tabular}[c]{@{}c@{}}79.15 ±\\ 2.25\end{tabular}} & \textbf{\begin{tabular}[c]{@{}c@{}}83.0 ±\\ 1.7\end{tabular}} & \begin{tabular}[c]{@{}c@{}}84.25 ±\\ 1.55\end{tabular} \\
 & Llama2-7B (Generalization) & - & - & - & \begin{tabular}[c]{@{}c@{}}{\ul 22.75} ±\\ {\ul 0.65}\end{tabular} & \begin{tabular}[c]{@{}c@{}}36.2 ±\\ 0.7\end{tabular} & \begin{tabular}[c]{@{}c@{}}44.0 ±\\ 0.8\end{tabular} & \begin{tabular}[c]{@{}c@{}}{\ul 13.75} ±\\ {\ul 0.95}\end{tabular} & \begin{tabular}[c]{@{}c@{}}20.35 ±\\ 1.05\end{tabular} & \begin{tabular}[c]{@{}c@{}}27.6 ±\\ 0.8\end{tabular} & \begin{tabular}[c]{@{}c@{}}68.9 ±\\ 0.6\end{tabular} & \begin{tabular}[c]{@{}c@{}}75.45 ±\\ 0.35\end{tabular} & \begin{tabular}[c]{@{}c@{}}82.05 ±\\ 0.35\end{tabular}\\
 \bottomrule
\end{tabular}
}\end{center}
\end{table*}

\subsection{Main Results}
To answer the RQ1, our results from Table \ref{tab: link prediction results PART1} achieve state-of-the-art performance, surpassing all three types of existing conventional including embedding-based models, rule-based method, and LLM-based in-context learning method across four datasets regarding metric Hit@1 and Hit@3 while maintaining comparable results regarding Hits@10. Our method demonstrates the promising trend for retrieval-augmented LLMs to serve as the foundation model for temporal relational forecasting, opening a new frontier in the TKG domain. We refer to GenTKG utilizing LLaMA2-7B as instantiation unless otherwise specified.  \\
\textbf{Compared to embedding-based methods.} For all datasets, GenTKG outperforms its best embedding-based model xERTE on ICEWS14, ICEWS18, GDELT, and Timetraveler on YAGO. Specifically, the highest performance gain is observed on GDELT with more than 58\% higher on Hits@1.\\% It is natural to conclude that GenTKG can outperform embedding-based methods. \\
\textbf{Compared to the rule-based method.} Compared to the rule-based model TLogic, GenTKG outperforms TLogic on Hits@1 and Hits@3 while maintaining comparable performance regarding Hits@10. The slight drops regarding Hits@10 on ICEWS14 and ICEWS18 are because TLogic is carefully designed on these datasets while our method has more generalizability and demonstrated better performance regarding accuracy than recall. \\
%with relatively 12\% higher Hits@1 and 7\% higher Hits@3. For dataset, ICEWS18, GenTKG outperforms its best embedding-based model Timetraveler with relatively 2\% higher Hits@1 and 14\% higher Hits@3. Regarding Hits@10, GenTKG achieves comparable performance with different embedding-based models. It is naturally concluded that GenTKG has overall better performance than embedding-based models.
% \textbf{Compared to rule-based model}
% Similarly, GenTKG outperforms TLogic regarding both Hits@1 and Hits@3, with 12\% and 2\% on ICEWS14, while 6\% and 10\% on ICEWS18. However, performance drops on both datasets regarding Hits@10 are witnessed which implies that GenTKG has a better capability of forecasting hard negative samples with higher accuracy while TLogic has better recall performance. The difference is likely attributed to the short history retrieval length due to the limited context window size of LLM, which we will discuss more in the later Limitation section \ref{}. 
\textbf{Compared to in-context-learning method.}
We analyze the performance of GenTKG on different Language Model instantiations, i.e. GPT-NeoX-20B and LLaMA2-7B respectively. For GPT-NeoX-20B, we apply only the first retrieval phase of GenTKG due to hardware limitations. However, a huge performance increase is observed for all three metrics on all datasets even with pure retrieval-augmented in-context learning. For LLaMA2-7B, the performance gain of Hits@1 has increased remarkably even outperforming GPT-NeoX-20B which has two times more parameters, indicating the potential for greater performance of our proposed GenTKG framework if applied to larger language models. %specifically, 9\% higher Hits@1, 10\% higher Hits@3 and 10\% higher Hits@10 on ICEWS14 and 9\% higher Hits@1, 10\% higher Hits@3 and 10\% higher Hits@10 on ICEWS14

\subsection{Cross-domain Generalization}\label{sec: cross-d ind}
\begin{figure}
    \centering
    \includegraphics[width=1.0\linewidth]{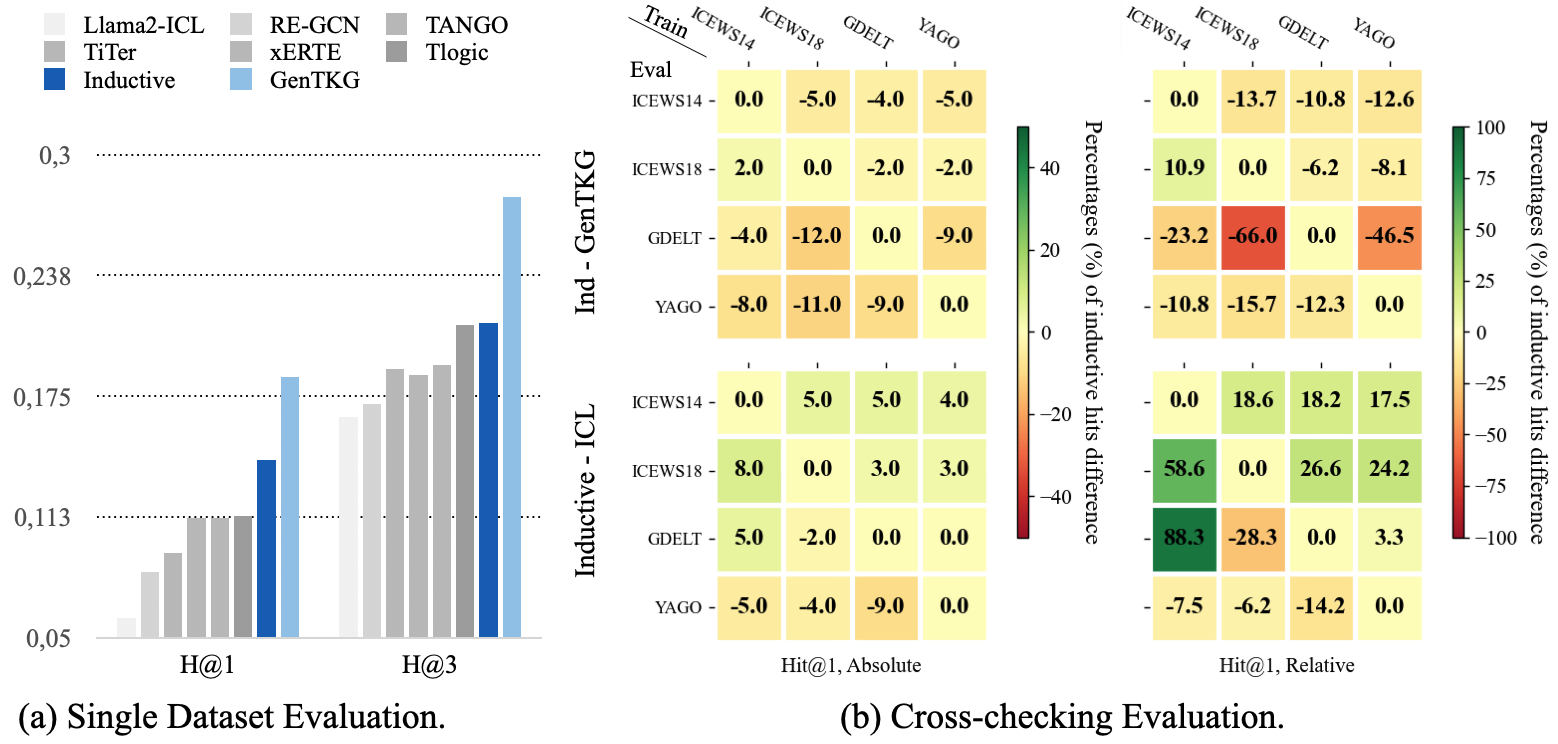}
    \caption{Cross-Domain Generalization Setting. (a) Single dataset evaluation. All training and evaluation is on GDELT except generalized GenTKG, which is trained on ICEWS14. (b) Cross-checking. We cross-check the trained LLaMA2 in GenTKG on different training datasets and evaluation datasets. The performance drop compared to the original training setting takes up only small percentages.  Even higher performance than ICL can be observed. 
    % Absolute difference value 
    More discussions about experiment settings and analysis are given in Appendix \ref{sss: ap4}, explaining the huge relative difference on GDELT is due to its poor baseline performances.}
    \label{fig:ind-cross}
\end{figure}

To answer the second question of GenTKG's performance in the generalization setting, the empirical results indicate that the GenTKG framework manifests a substantial capability for cross-dataset generalization. Specifically, once the LLM has been aligned to the tKG forecasting task in the second phase on any dataset, the LLM can be applied directly to any other dataset. Therefore, on a new dataset, GenTKG only requires dataset-specific temporal-logical rule-based retrieval to formulate proper prompts from the first phase, and can directly infer the predictions without retraining in the second phase. As shown in Figure \ref{fig:ind-cross}(a), all methods are trained and evaluated on GDELT, except that the LLM in generalized GenTKG is trained ICEWS14. Still, the generalized GenTKG delivers comparable performance metrics on GDELT to conventional methods with a minor performance drop compared to the original trained GenTKG. We further demonstrate similar generalization results by cross-checking the training and evaluation datasets as shown in Figure \ref{fig:ind-cross}(b). Although the LLM is trained exclusively on one dataset, it still delivers comparable metrics on disparate datasets, closely approximating the outcomes of methods that were trained specifically on the identical evaluation dataset. This notable characteristic implies that the GenTKG is effectively capturing the underlying task-related features, as opposed to merely carefully designed for the dataset data, a limitation commonly shared in conventional methods.

% \begin{figure}
%     \centering
%     \includegraphics[width=1.0\linewidth]{figures/ind-full.png}
%     \caption{Cross-Domain Inductive Setting. (a) All models including GenTKG are trained and evaluated on the GDELT dataset, except that the inductive setting of GenTKG is trained on ICEWS14 and evaluated on GDELT.  (b) Cross-checking. We cross-check the trained LLaMA2 in GenTKG on different training datasets and evaluation datasets. The performance drop compared to the original training setting takes up only small percentages.  Even higher performance than ICL can be observed.}
%     \label{fig:ind-cross}
% \end{figure}

\subsection{In-domain Generalization}
\begin{figure}
    \centering
    \includegraphics[width=1.0\linewidth]{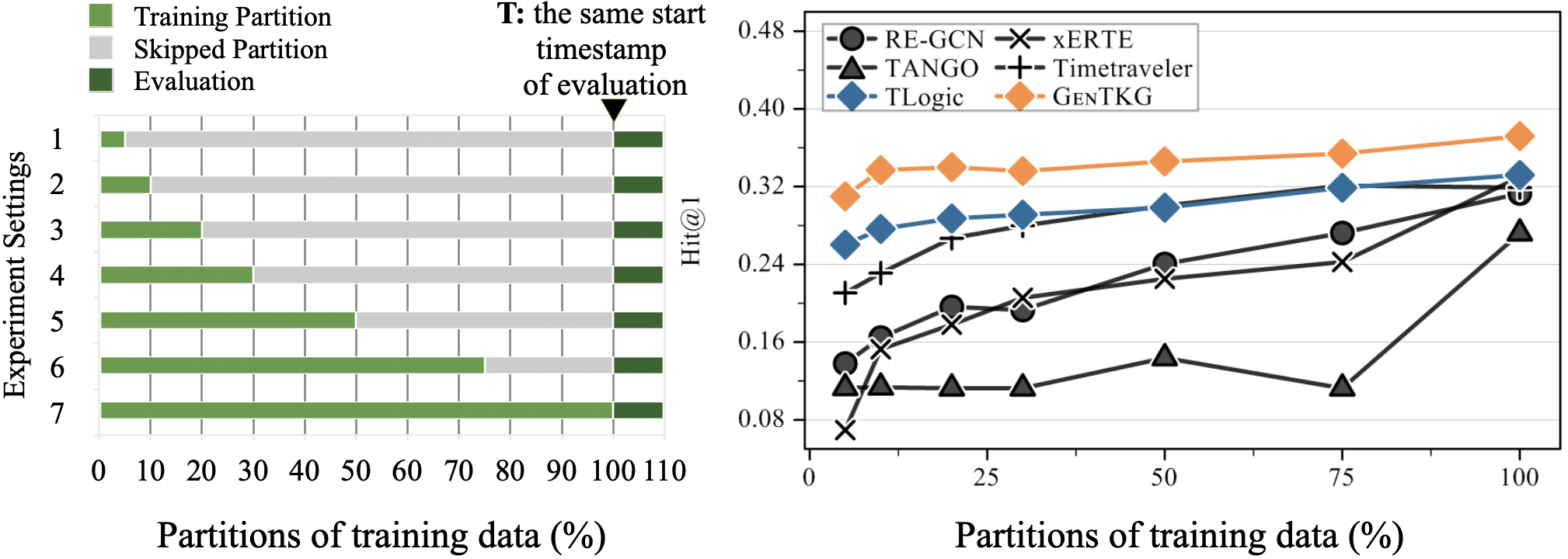}
    \caption{In-domain generalizability. GenTKG exceeds conventional methods on all different partitions of training data on ICEWS14. Values in Appendix Table \ref{tab:few-shot-ind}.}
    \label{fig:ind-in}
\end{figure}

Apart from cross-domain generalizability, how well does GenTKG generalize to different training partitions within the same dataset? To investigate such a problem, we carefully designed various partitions of time-ordered training data ranging in \{5\%, 10\%, 20\%, 30\%, 50\%, 75\%, 100\%\}. All models trained on different training partitions are evaluated on the same evaluation set starting from the same timestamp. According to Figure \ref{fig:ind-in}, experiments have shown that conventional methods suffer from insufficient training data while GenTKG remains exceeding performance even with as few as 5\% training data. This further proves that GenTKG successfully transforms conventional data-centric learning to the task-centric alignment of LLMs and overcomes the prediction instability under the changing value of time split in the forecasting setting.

\subsection{Ablation study}
\begin{figure}
    \centering
    \includegraphics[width=0.9\linewidth]{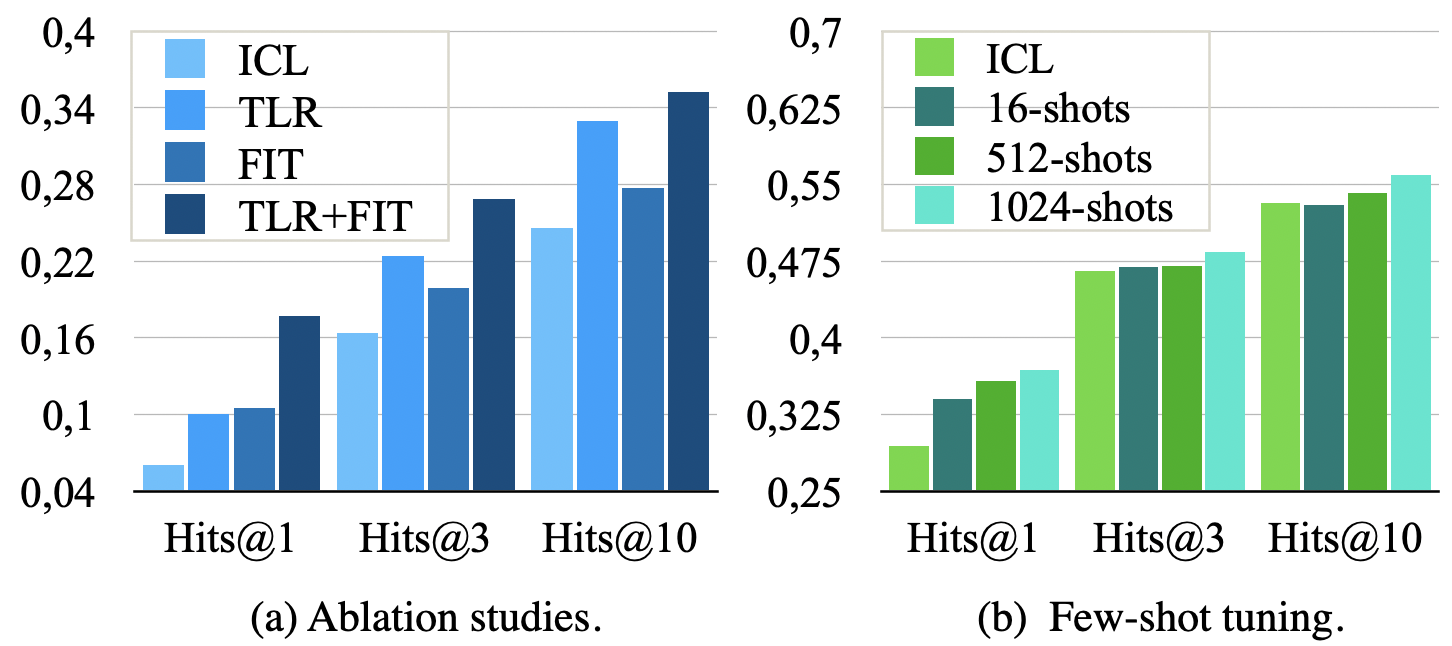}
    \caption{(a) Both TLR and FIT phases contribute to GenTKG. (b) Increasing the few-shot training parameter $K$ improves performance.}
    \label{fig:ab studies}
\end{figure}

%To comprehensively evaluate the contribution of each constituent element within our proposed GenTKG framework
We undertake ablation studies on ICEWS14 to evaluate the contribution of each phase in GenTKG with three distinct variants of the GenTKG: TLR, FIT, and TLR+FIT configurations. Here, TLR represents the variant that exclusively employs temporal logical rule-based retrieval on top of ICL learning, FIT denotes the variant solely implementing few-shot parameter-efficient instruction tuning with naive fact retrieval~\citep{lee2023temporal}, and TLR+FIT encapsulates the integration of all components within GenTKG. Figure \ref{fig:ab studies}(a) draws the conclusion that both phases in GenTKG framework contribute to distinct performance improvements. The whole pipeline enables GenTKG the ability to outperform existing methods.

% \subsection{Inductive Setting}

% \begin{table}
%     \caption{Caption}
%     \centering
%     \begin{tabular}{ccccccc}
%          Few-shots& Hits@1 & Hits@3 & Hits@10 &  &  & \\
%          16&  &  &  &  &  & \\
%          512&  &  &  &  &  & \\
%          1024&  &  &  &  &  & \\
%     \end{tabular}
%     \label{tab:my_label}
% \end{table}
% \section{Discussions}

% \textbf{Backbone Language Models}

\subsection{Few-shot Tuning}
To delve further into the impact of sample size within the few-shot tuning, we conducted a series of experiments on the ICEWS14 dataset employing a range of shot sizes $K$ from the set $\{16, 512, 1024\}$. For each configuration, we employed uniform sampling on the temporally-ordered training dataset. Empirical results indicate a consistent trend of performance improvement correlating proportional to the increase in the number of training samples, as visualized in Figure \ref{fig:ab studies}(b). Remarkably, our findings suggest that the GenTKG framework is capable of outperforming naive ICL method even when as few as 16 shots are used for tuning. This notable finding unlocks significant potential for GenTKG in the context of aligning LLMs with temporal relational forecasting tasks from the perspective of efficient alignment or a larger scale.

% \textbf{History length Limit}

\section{Discussion}
% \subsection{Q1:How does History Length Affect GenTKG?}

\paragraph{Q1: How does the index or lexical format of the prompt affect the results?}

% First of all, we would like to emphasize the reason why we opted to choose the lexical format. Since [1] is a respectful baseline of our work, we would like to compare the results fairly. In the work of [1], LLaMA2 is not reported due to two reasons. The first is the LLaMA-7B’s poor in-context learning ability compared to GPT-NeoX-20B model. This might originate from the models’ parameter size difference, which, however, is flipped with GenTKG with a successful retrieval strategy and a well-designed few-shot tuning strategy. The second reason is the tokenization difference between the two model families. GPT-series models harness tokenization that treats the index of an entity as a whole no matter the digit length of such index, while LLaMA-series models harness individual digit tokenization. Consequently, the probability of an index is calculated by LLaMA2 as the product of probabilities of each digit hence causing unfairness to index with different lengths since LLaMA2 will always bias towards shorter index as prediction. This technical unfairness contradicts the intent of reasoning over the temporal relational forecasting task itself. Therefore, to conduct a fair comparison to [1] we opt to utilize a lexical format for evaluation.

To ease the concern of data leaks in the pre-training process of LLMs, we investigate whether the lexical or index format prompt affects the LLM generative forecasting ability. We conduct experiments with ChatGPT~\footnote{ChatGPT (gpt-3.5-turbo) version 02.2024 is used here.} using index format following ICL baseline settings in ~\cite{lee2023temporal} as a fair comparison. Due to the restriction of training ChatGPT, we equipped ChatGPT with the temporal logical retrieval strategy (TLR) of GenTKG compared to the ICL baseline in both lexical and index form. The experiment results are reported in Table \ref{tab: id_vs_lex}.

 \begin{table}[!ht]
 \caption{Performance (Hits@1) between index and
lexical for gpt-3.5-turbo on ICEWS14. }
 \label{tab: id_vs_lex}
    \centering
    \resizebox{0.45\textwidth}{!}{
    \begin{tabular}{c|c|c|c}
    \hline
        \multirow{2}{*}{Configuration} & \multirow{2}{*}{Model} & \textbf{lexical} & \textbf{index} \\
         & & Hits@1 &Hits@1 \\ \hline
        GenTKG-TLR& gpt-3.5-turbo& 0.21& 0.26\\ \hline
         ICL& gpt-3.5-turbo & 0.18 & 0.16 \\ \hline
    \end{tabular}
}
\end{table}

Three interesting insights can be derived here. (1) First, the index form conter-intuitively outperforms the lexical form and therefore the concern of data leakage in the pre-trained LLMs is relieved. (2) Second, our TLR retrieval strategy steadily outperforms ICL baseline retrieval on ChatGPT, further proving its LLM-agnostic retrieval enhancement. (3) 
%The third is more interesting and inconsistent with the ICL baseline. ~\cite{lee2023temporal} finds that foundation models depend more on input-label mappings and learning the pattern and tKG forecasting tasks are impacted to a small extent by semantic priors although 
Instruction-tuned models like ChatGPT should make better use of semantic priors. However, our reverse results in the configuration of GenTKG-TLR indicate that the successful TLR retrieval strategy, which heavily relies on the temporal and structural patterns, lets instruction-tuned models like ChatGPT grasp latent patterns more easily with index and outweigh the benefit brought by semantic priors. This reveals the ability of LLM to learn temporal relational patterns more than relying on semantic priors, which we believe is a beneficial finding for future research.

\paragraph{Q2: How well is the qualitive improvement of TLR retrieved facts?}
We conduct a qualitative study regarding the temporal logic rule-based retrieval strategy (TLR) to intuitively understand its retrieval quality. 
% There’s a common phenomena comparing the TLR retrieved history and ICL-baseline~\cite{lee2023temporal}. 
The ICL-baseline~\cite{lee2023temporal} retrieves the most recent histories and retrieves histories igoring relation relevance. While TLR retrieves history with temporal logic rules and therefore the relations in the history facts are more related to the query. For example, given the query \shortstack{\textit{334: [Abdul, Make\_an\_appeal\_or\_request,? ]}}, ICL-baseline retrieves facts mostly with general relations like \textit{Host\_a\_visit} and \textit{Make\_a\_visit}. However, TLR retrieves facts containing relations like \textit{Acknowledge\_or\_claim\_responsibility} and \textit{Cooperate\_militarily}, which are significantly more logically relevant. These two respective rules are visible in the TLR rule bank with high confidence, which justifies the better predictive performance with precise retrieval.

\paragraph{Q3: How does temporal information affect GenTKG?}
To assess how GenTKG comprehend the temporal information of historical events, we set four temporal configurations on ICEWS14. \textit{Original} organizes retrieved facts into ascending order, where the latest event is set closest to the test query, while \textit{Reverse} configuration is in descending order. We further set two settings with \textit{Random} temporal order and \textit{Removal} of timestamp.

%  \begin{table}[!ht]
%  \caption{Performance of different temporal settings in prompts on ICEWS14. }
%  \label{tab: abl_temporal}
%     \centering
%     \resizebox{0.35\textwidth}{!}{
%     \begin{tabular}{c|c|c|c}
%     \hline
%         Configuration & Hits@1 & Hits@3 & Hits@10 \\\hline
%         Original & 0.369 & 0.480 & 0.535 \\ \hline
%         Reverse & 0.152 & 0.342 & 0.487 \\ \hline
%         Random & 0.246 & 0.403 & 0.493 \\ \hline
%         Removal & 0.280 & 0.411 & 0.484 \\ \hline
%     \end{tabular}
% }
% \end{table}

\begin{figure}
    \centering
    \includegraphics[width=1.0\linewidth]{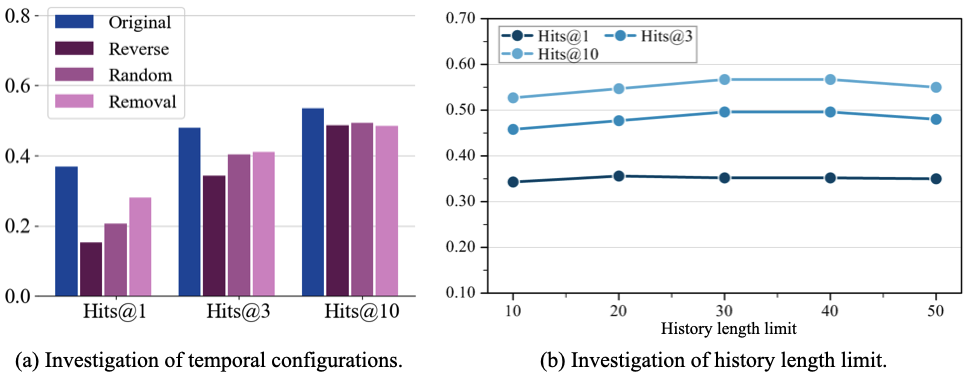}
    \caption{(a) Other temporal configurations deteriorate performance. (b) Increasing the history length limit improves performance.}
    \label{fig: dis}
\end{figure}

The results in Figure \ref{fig: dis}(a) show that all configurations other than the original ascending order lead to a deterioration in performance. Among them, the \textit{Removal} indicates a least performance deterioration implying that the sequential order of events has an implicit consistency in the \textit{Original} ascending order for LLM to reason the temporal information.

\paragraph{Q4: How does history length affect GenTKG’s performance?}
Due to the limitation of LLM context length, we evaluate the impact of the history
length of TLR retrieved facts. We conduct a set of experiments on ICEWS14 using varying truncated history lengths, i.e. the upper length limit, with four configurations $\{10,20,30,40,50\}$. Our
results, as shown in Figure \ref{fig: dis}(b), indicate that improving history length generally leads to better performance and imply that most temporal and logically relevant facts are retrieved in the near past, and retrieving less relevant facts will affect performance.

% a consistent
% improvement in performance as the history length
% increases. This suggests that the models learn better as additional historical facts are presented. This
% observation is connected to few-shot learning in
% other domains, where performance improves as the
% number of examples per label increases. However,
% in our case, the historical patterns presented in the
% prompt do not explicitly depict the input-label mapping but rather aid in inferring the next step.

\section{Conclusion}
This paper raises the question and proves that pre-trained LLMs can understand structured temporal relational data and replace existing tKG models as the foundation model for temporal relational forecasting task.  %We then attempt to adapt LLM through retrieval-augmented generation approach GenTKG and investigate the feasibility of using LLM for the temporal relational forecasting.  
We propose a retrieval-augmented generative framework GenTKG that can efficiently align LLM with temporal relational forecasting task through two stages: temporal logical rule-based retrieval (TLR) and few-shot parameter-efficient fine-tuning (FIT).
Extensive experimental results demonstrate that GenTKG outperforms conventional embedding-based, rule-based and ICL methods. Moreover, GenTKG is training-light through comsumable computation resources with extremely few training data, and exhibits strong cross-domain and in-domain transferability breaking the barriers of conventional data-centric learning. %We further reveal the potential of GenTKG as a model-agnostic framework for various LLMs and open a new strong frontier for utilizing LLMs as foundation models for solving temporal relational forecasting tasks.

\section{Limitation and Future Directions}
% Although our method significantly outperforms the existing approaches in various types, our approach is limited by the inference time of large language models. Specifically, for LLaMA2, the inference time for a simple query is 6 second in GenTKG, which limits large-scale inference on tKG. We leave this to future work.
GenTKG is limited by the input context window of LLMs. Specifically, for LLaMA2, the input context window is 4096 tokens with an average upper length limit of 50 history facts that limit the performance of Hit@10.
% It is likely that the relevant facts will not be fully retrieved from the first phase limiting the generative forecasting ability of the second phase in GenTKG. 
This RAG framework of GenTKG has the potential to combine better retrieval strategies and prompt LLMs with longer context windows to fully utilize temporal reasoning of LLMs. The strong generalization ability may also benefit inductive settings, zero-shot, or few-shot tasks in tKG, which we leave to the future.

\newpage
\section*{Ethics Statement}
% Scientific work published at EMNLP 2023 must comply with the \href{https://www.aclweb.org/portal/content/acl-code-ethics}{ACL Ethics Policy}. We encourage all authors to include an explicit ethics statement on the broader impact of the work, or other ethical considerations after the conclusion but before the references. The ethics statement will not count toward the page limit (8 pages for long, 4 pages for short papers).

% In terms of ethical considerations regarding the dataset, we implemented OpenAI moderation APIs\footnote{\url{https://platform.openai.com/docs/api-reference/moderations}} to screen for potentially harmful questions, including those involving violence, sexual content, or hate speech. The results revealed that no questions were flagged as containing harmful content.

% Regarding potential ethical concerns with the CrossGNN model, it utilizes both pre trained knowledge embedded in the model and the graph modality for text generation. Consequently, CrossGNN has the potential to manifest biases and incorporate toxic information present within knowledge graphs and pre trained language models.

GenTKG is tailored to generative forecasting on temporal knowledge graphs and can be applied to a wide variety of downstream tasks with generative forecasting settings, such as recommendation systems, anomaly detection, etc. It can also power search and serve to improve users’ lives. GenTKG can help protect data with its generalizability which requires less training over various datasets. The risk of GenTKG might come from risks inherited in open-source LLMs, such as hallucinations.

\section*{Liscence}
The datasets used in this research work are open-sourced and can be seen in references. We derive some datasets from the original version within the intended use term. For the GDELT dataset, as stated in the terms of use of
GDELT\footnote{https://www.gdeltproject.org/about.html\#termsofuse}, this project is an open
platform for research and analysis of global
society and thus all datasets released by the
GDELT Project are available for unlimited
and unrestricted use for any academic, commercial, or governmental use of any kind without fee. One may redistribute, rehost, republish, and mirror any of the GDELT datasets in any form. However, any use or redistribution of the data must include a citation to the GDELT Project and a link to this website (https://www.gdeltproject.org/). ICEWS follows the MIT license on its official website (https://github.com/andybega/icews?tab=MIT-2-ov-file) and YAGO is licensed under a Creative Commons Attribution 4.0 International License (https://yago-knowledge.org/).

\section*{Acknowledgements}

This work was funded by the Munich Center for Machine Learning and supported by the Federal Ministry of Education and Research and the State of Bavaria. We are thankful for the helpful discussions with Yushan Liu from Siemens AG and Zifeng Ding from LMU Munich. 

\appendix

\bibliographystyle{IEEEtran}
\bibliography{reference}

\newpage
\appendix

% \section{Examples of GraphextQA Dataset}

% % \begin{table*}
% % \centering
% % \begin{tabular}{lll}
% % \hline
% % \textbf{Question} & \textbf{Answers} & \textbf{Subgraph}\\
% % \hline
% % AAA & BBB & CCC \\

% % \hline
% % \end{tabular}
% % \caption{\label{more-examples}
% % Random examples from the proposed GraphextQA datasets.
% % }
% % \end{table*}

% \begin{table*}[htbp]
%   \centering
%   \begin{tabular}{p{5cm}|l|p{8cm}}
%     \hline
%     \textbf{Question} & \textbf{Answers} & \textbf{Subgraph} \\
%     \hline
%     % What American person did The Dictator and This Must Be the Place star & \[ "Barack Obama" \] & \texttt{\{ "entities": \[ "Q30", "Q5", "Q76", "Q511347", "Q464964" \], "relations": \[ "P27", "P31", "P161" \], "adjacency": \[ \[ 2, 1, 1 \], \[ 4, 2, 2 \], \[ 3, 2, 2 \], \[ 2, 0, 0 \] \], "entity_labels": \[ "United States of America", "human", "Barack Obama", "This Must Be the Place", "The Dictator" \], "relation_labels": \[ "country of citizenship", "instance of", "cast member" \] \}} \\
%     % \hline
%     Who advised Benedict XVI on his doctorate? & \[ "Hub Schnackers" \] & \texttt{\{ "entities": \[ "Q5", "Q2494", "Q46378450" \], "relations": \[ "P184", "P31" \], "adjacency": \[ \[ 2, 0, 1 \], \[ 2, 1, 0 \] \], "entity_labels": \[ "human", "Benedict XVI", "Hub Schnackers" \], "relation_labels": \[ "doctoral advisor", "instance of" \] \}} \\
%     \hline
%   \end{tabular}
%   \caption{Random examples from the proposed GraphextQA datasets.}
%   \label{tab:more-examples}
% \end{table*}

\section{Related Works} \label{sec:appendix}
\paragraph{Temporal Knowledge Graphs}
Temporal knowledge graphs (tKGs) are multi-relational, directed graphs with labeled timestamped edges between entities (nodes). Let $\mathcal{E}$ and $\mathcal{P}$ represent a finite set of entities and predicates. A quadruple $(e_s, r, e_o, t)$ represents a timestamped and labeled edge between a subject entity $e_s \in \mathcal{E}$ and an object entity $e_o \in \mathcal{E}$ at a timestamp $t \in \mathcal{T}$ . Let $\mathcal{F}$ represent the set of all true quadruples, i.e., real
events in the world, the temporal knowledge graph forecasting task predicts missing object entity at timestamp $t$, i.e. $(e_s, r, ?, t)$ based on a set of observed facts $\mathcal{O}$ before $t$, which is a subset of $\mathcal{F}$.  Current methods can be categorized into two streams. Embedding-based models learn representations of the quadruples with carefully designed embedding models~\cite{han2020explainable, goel2020diachronic, sun2021timetraveler, han2020dyernie, ding2022few}. Rule-based methods mine the temporal logical rules extracted and extract candidates directly on the tKGs~\cite{liu2022tlogic}. 

\paragraph{Investigating Static KG with LLMs}
Later ideas also investigated static KG with LLMs utilizing the knowledge-aware prompting methods~\cite{galkin2023towards, li2024condensed, baek-etal-2023-knowledge, rony2022dialokg,sun2023think,zhang2022drlk}. However, they cannot be transferred to the tKG domain due to their ignorance of temporal characteristics. Specifically, ~\citep{li2024condensed} uses structured retrieved triples for reasoning on KG and conducts a much simpler task of reasoning on static KG. GenTKG is not only more pioneer but also more powerful since tKG forecasting is more difficult due to its temporal dynamics and we contribute our distinct RAG framework for temporal reasoning with LLMs.

\paragraph{Investigating TKG with Language Models}
The semantic part stored in the temporal knowledge graphs is heavily overlooked in either embedding-based or rule-based temporal knowledge graph methods. Early explorers had tryouts in introducing language models in the TKG domain, some fused pre-trained language representations to the temporal knowledge embeddings~\citep{han2022enhanced}, and some flattened explicit temporal events with the emergent in-context learning ability of large language models however not comparable with conventional performance~\citep{lee2023temporal}. ~\citep{ding2023zero} explores LLM in the zero-shot relational learning settings in the TKG forecasting task.

\newpage
\section{Supplimentary Materials}\label{sec:supplimentary}

\subsection{Discussion on Cross-domain Generalizability} \label{sss: ap4}
We give further details regarding cross-domain generalizability experiments in Sec \ref{sec: cross-d ind} and Figure \ref{fig:ind-cross}.
\paragraph{Cross-check Experiment Settings}
To assess the cross-domain generalizability according to the 4 test benchmarks in this paper, we conduct 4 series of cross-domain generalization settings respective to each benchmark. We define that a series of cross-checking settings consists of a center evaluation dataset $\mathcal{A}$ with the other three cross-checking datasets denoted as $\mathcal{B, C, D}$. Inside a series, a single evaluation on $\mathcal{A}$ is conducted by comparing all inference results on the center $\mathcal{A}$ including (1) all baseline methods trained on $\mathcal{A}$, (2) original GenTKG trained on $\mathcal{A}$, and (3) generalized GenTKG trained one of the other three cross-checking dataset, e.g. $\mathcal{B}$. In total, $4 \times 3=16$ cross-checking experiments are conducted.

% We further visualize each series by comparing all baseline methods including GenTKG trained and evaluated on dataset $\mathcal{A}$ with the inductive GenTKG trained on datasets $\mathcal{B, C, D}$.

% The definitions of the inductive settings in our work. Overall, the inductive setting is on the scale of “dataset” instead of “nodes”. The cross-domain inductive setting means we train on the single dataset and inference the model directly on new datasets. Hence, all nodes in the new datasets are new nodes and not seen by the model.

\paragraph{Experiment Results}
Figure \ref{fig:ind-cross} reports results for the cross-domain generalizability of GenTKG. 
Figure \ref{fig:ind-cross}(a) visualizes a single evaluation in a series, taking GDELT as the evaluation dataset, and ICEWS14 as the cross-checking dataset for an example. Figure \ref{fig:ind-cross}(b) visualizes the result differences between generalized GenTKG compared with original GenTKG (-Ori), and compared with ICL baseline (-ICL). The upper row represents the relative difference($\%$) of generalized GenTKG subtracted by original GenTKG (-Ori). The lower row represents the relative difference($\%$) of generalized GenTKG subtracted by ICL baseline (-ICL). Please refer to Table \ref{tab:absdiff} with absolute value differences and Table \ref{tab:reldiff} with relative value differences. 

Regarding Fig \ref{fig:ind-cross}(a), similar patterns can be seen in other series of cross-checking. Table \ref{tab: link prediction results PART1} with the last row reports the results of GenTKG trained on ICEWS14 and tested with on other 3 datasets ICEWS18, GDELT, and YAGO. The generalized GenTKG trained on ICEWS14 have comparable and even exceeding results on ICEWS18, GDELT, and YAGO, compared to baselines, and suffer only slight drops compared to the setting of original training setting.

Regarding Fig \ref{fig:ind-cross}(b), two conclusions can be drawn. First, generalized settings have a performance drop compared to the original ones. Second, similar datasets tend to have better generalization performance when exchanging the training dataset. Third, even the cross-checking setting of the distant dataset can obtain better performance than ICL with minor cases of performance drop but still comparable.

This is accountable for dataset similarities. ICEWS14 and ICEWS18 originate from the same political event database with differences in the year where the data come from. ICEW14 collects data from 2014 while ICEWS18 from 2018 with a time interval of day. Therefore, the two datasets share similar patterns regarding events and patterns. GDELT documents events between global entities with a time interval of 15 minutes and YAGO originates from WIKI with a time interval of year hence they contain more complex relations and are much more distant than that between ICEWS14 and ICEWS18.

\subsection{Detailed TLR Algorithm} \label{sss: ap_alg}

\begin{algorithm}
\SetAlgoLined
\Input{Temporal knowledge graph $\mathcal{G}$, query $(e_s, r, ?, t)$}
\Parameter{Time window length $w \in \mathbb{N}^+$, learned rules $\mathcal{TR}$}
\Output{A set of retrieved facts $\mathcal{G}_s(e_s, r, t)$}

 $\mathcal{G}_s(e_s, r, t)\leftarrow\emptyset$\;
 \For{$(e_s, r, ?, t) \in \mathcal{G}$}{
  $TW \leftarrow[t-w,t]$\;
  \For{$fact\leftarrow (e_s, r_{h}, e_o, t-w<t) \in \mathcal{G}$}{
    $\mathcal{G}_s(e_s, r, t)\leftarrow\mathcal{G}_s(e_s, r, t)\cup fact$
  }
  \For{top $k$ rules w.r.t $r_h \leftarrow r_b$ $\in \mathcal{TR}$}{
    Get a list $r_{b}\leftarrow{r_{b_1}, r_{b_2}, \cdots}, r_{b_k}$
  }
  \For{$fact\leftarrow (e_s, r\in r_{b}, e_o, t-w<t) \in \mathcal{G}$}{
    $\mathcal{G}_s(e_s, r, t)\leftarrow\mathcal{G}_s(e_s, r, t)\cup fact$
  }
  %$TW \leftarrow[t_{-},t]$不知道怎么用
\Return{$\mathcal{G}_s(e_s, r, t)$}
\caption{TLR Retrieval}
}
\end{algorithm}

\pagebreak

\subsection{Implementation Details}
We run experiments 3 times and take averages with A40 GPU. For the TLR part, we use the rule length of 1, the number of random walks of 200, the time window of the maximum length of each dataset, and the maximum history length of 50. In the FIT training, we use the batch size of 1024, the learning rate of $3e-4$, the context length of 4096, the target length of 128, the LoRA rank of 8, the LoRA dropout rate of 0.05, and few-shot tuning of 1024-shots. Besides, we use the Adam optimizer~\citep{kingma2014adam}.

\subsection{Supplementary Statistics} \label{sss: ap_dataset stat}

 \begin{table}[!ht]
 \caption{Dataset statistics.}
 \begin{center}
\resizebox{0.5\textwidth}{!}{
 \label{tab: dataset}
    \centering
    \resizebox{0.6\textwidth}{!}{
    \begin{tabular}{l|l|l|l|l|l|l}
    \hline
        Datasets & \#train & \#valid & \#test & \#entity & \#relations & time gap \\ \hline
        ICEWS14 & 74854 & 8514 & 7371 & 7128 & 230 & 1 day \\ \hline
        ICEWS18 & 373018 & 45995 & 49545 & 23033 & 256 & 1 day \\ \hline
        GDELT & 79319 & 9957 & 9715 & 5850 & 238 & 15 mins \\ \hline
        YAGO & 220393 & 28948 & 22765 & 10778 & 23 & 1 year \\ \hline
    \end{tabular}
}}\end{center}
\end{table}

% \begin{table*}[htbp]
\begin{table*}[h]
\caption{Absolute difference value for cross-checking between generalized GenTKG compared with original GenTKG (-ori), and compared with ICL baseline(-ICL).
}
\label{tab:absdiff}
\begin{center}
\resizebox{1.0\textwidth}{!}{
% Please add the following required packages to your document preamble:
% \usepackage{multirow}
\begin{tabular}{l|l|llll|llll|llll}
\toprule
\multirow{2}{*}{} & \multirow{2}{*}{\diagbox{Eval}{Train}} & \multicolumn{4}{l|}{Hits@1} & \multicolumn{4}{l|}{Hits@3} & \multicolumn{4}{l}{Hits@10} \\
 &  & ICEWS14 & ICEWS18 & GDELT & YAGO & ICEWS14 & ICEWS18 & GDELT & YAGO & ICEWS14 & ICEWS18 & GDELT & YAGO \\ \midrule
\multirow{4}{*}{$\Delta$(-Ori)} & ICEWS14 & - & -0.05 & -0.04 & -0.05 & - & -0.05 & -0.03 & -0.03 & - & -0.04 & -0.05 & -0.05 \\
 & ICEWS18 & 0.02 & - & -0.02 & -0.02 & -0.02 & - & -0.02 & -0.02 & -0.02 & - & -0.04 & -0.04 \\
 & GDELT & -0.04 & -0.12 & - & -0.09 & -0.07 & -0.15 & - & -0.10 & -0.08 & -0.17 & - & -0.11 \\
 & YAGO & -0.08 & -0.11 & -0.09 & - & -0.07 & -0.09 & -0.06 & - & -0.02 & -0.06 & -0.03 & - \\ \midrule
\multirow{4}{*}{$\Delta$(-ICL)} & ICEWS14 & - & 0.05 & 0.05 & 0.04 & - & -0.01 & -0.01 & 0.00 & - & -0.02 & -0.03 & 0.01 \\
 & ICEWS18 & 0.08 & - & 0.03 & 0.03 & 0.04 & - & 0.02 & 0.02 & 0.05 & - & 0.04 & 0.07 \\
 & GDELT & 0.05 & -0.02 & - & 0.00 & 0.04 & -0.08 & - & -0.03 & 0.03 & -0.11 & - & 0.00 \\
 & YAGO & -0.05 & -0.04 & -0.09 & - & -0.05 & -0.04 & -0.09 & - & -0.09 & -0.07 & -0.10 & - \\ 
\bottomrule
\end{tabular}
}
\end{center}
% \end{table*}

\bigskip

% \begin{table*}[htbp]
\caption{Relative difference (\%) value for cross-checking between generalized GenTKG compared with original GenTKG (-ori), and compared with ICL baseline(-ICL).
}
\label{tab:reldiff}
\begin{center}
\resizebox{1.0\textwidth}{!}{
% Please add the following required packages to your document preamble:
% \usepackage{multirow}
\begin{tabular}{l|l|llll|llll|llll}
\toprule
\multirow{2}{*}{} & \multirow{2}{*}{\diagbox{Eval}{Train}} & \multicolumn{4}{l|}{Hits@1} & \multicolumn{4}{l|}{Hits@3} & \multicolumn{4}{l}{Hits@10} \\
 &  & ICEWS14 & ICEWS18 & GDELT & YAGO & ICEWS14 & ICEWS18 & GDELT & YAGO & ICEWS14 & ICEWS18 & GDELT & YAGO \\ \midrule
\multirow{4}{*}{$\frac{\Delta(-Ori)}{Ori}\times100\%$} & ICEWS14 & - & -13.71 & -10.75 & -12.63 & - & -9.43 & -5.33 & -6.35 & - & -7.10 & -9.06 & -8.88\\
 & ICEWS18 & 7.83 &  - & -8.76 & -10.60 & -4.31 & - & -5.17 & -6.61 & -5.10 & - & -9.28 & -8.12\\
 & GDELT &  -23.24 & -65.95 & - & -46.49 & -23.38 & -53.96 & - & -35.25 & -21.51 & -46.65 & - & -31.01\\ 
 & YAGO & -10.77 & -15.66 & -12.31 & - & -9.13 & -12.13 & -7.69 & - & -3.05 & -7.51 & -4.20 & -\\  \midrule
\multirow{4}{*}{$\frac{\Delta(-ICL)}{ICL}\times100\%$} & ICEWS14 & - & 18.65 & 18.25 & 17.46 & - & -3.28 & -1.87 & -0.94 & - & -2.98 & -4.96 & 1.79\\ 
 & ICEWS18 &  58.59 & - & 26.56 & 24.22 & 13.24 & - & 6.25 &  8.82 &  15.79 & - & 11.76 & 22.60\\ 
 & GDELT & 88.33 & -28.33 & - & 3.33 & 21.95 & -46.34 & - & -18.90 & 13.01 & -44.31 & - & 0.41\\ 
 & YAGO & -7.55 & -6.19 & -14.20 & - & -6.97 & -4.74 & -11.58 & - & -11.12 & -8.68 & -12.35 & -\\
\bottomrule
\end{tabular}
}\end{center}
% \end{table*}

\bigskip

% \begin{table*}[htbp]
\caption{Appendix table for few-shot results of conventional methods and GenTKG.
} \label{tab:few-shot-ind}
\begin{center}
\resizebox{1.0\textwidth}{!}{
\begin{tabular}{c|ccc|ccc|ccc|ccc|ccc|ccc|ccc}
\toprule
& \multicolumn{3}{c|}{Top 5\%}   & \multicolumn{3}{c|}{Top 10\%}  & \multicolumn{3}{c|}{Top 20\%}  & \multicolumn{3}{c|}{Top 30\%}  & \multicolumn{3}{c|}{Top 50\%}  & \multicolumn{3}{c|}{Top 75\%}  & \multicolumn{3}{c}{Top 100\%}  \\
& Hits@1& Hits@3& Hits@10   & Hits@1& Hits@3& Hits@10   & Hits@1& Hits@3& Hits@10   & Hits@1& Hits@3& Hits@10   & Hits@1& Hits@3& Hits@10   & Hits@1& Hits@3& Hits@10   & Hits@1& Hits@3& Hits@10   \\ 
\midrule
RE-GCN  & 13.79 & 22.09 & 30.27 & 16.47 & 25.23 & 34.19 & 19.63 & 29.67 & 39.83 & 19.30 & 30.66 & 42.97 & 24.05 & 36.72 & 48.84 & 27.23 & 40.42 & 54.04 & 31.30 & 47.30 & 62.60 \\
xERTE   & 06.95 & 14.17 & 25.46 & 15.27 & 26.79 & 39.43 & 17.80 & 29.26 & 42.08 & 20.56 & 31.39 & 43.63 & 22.51 & 34.15 & 46.59 & 24.25 & 36.07 & 48.27 & 33.00 & 45.40 & 57.00 \\
TANGO   & 11.29 & 17.18 & 22.97 & 11.34 & 17.47 & 22.98 & 11.25 & 17.38 & 23.38 & 11.25 & 17.39 & 23.40 & 14.37 & 17.51 & 22.77 & 11.25 & 16.90 & 22.50 & 27.20 & 40.80 & 55.00 \\
Timetraveler& 21.06 & 34.78 & 49.10 & 23.10 & 35.71 & 49.96 & 26.69 & 39.42 & 51.78 & 27.98 & 40.14 & 53.23 & 30.05 & 42.82 & 54.74 & 32.11 & 45.33 & 57.14 & 31.90 & 45.40 & 57.50 \\
TLogic Original & 26.03 & 37.42 & 46.50 & 27.65 & 39.55 & 48.72 & 28.72 & 40.48 & 50.71 & 29.11 & 41.79 & 51.90 & 29.84 & 42.40 & 53.37 & 31.89 & 45.01 & 57.37 & 33.20 & 47.60 & 60.20 \\ 
\midrule
GenTKG  & 30.60 & 42.20 & 49.30 & 34.00 & 45.40 & 52.10 & 34.90 & 46.60 & 54.00 & 34.70 & 46.90 & 54.40 & 36.00 & 48.70 & 55.50 & 36.50 & 48.30 & 55.30 & 37.20 & 48.80 & 56.30 \\ 
\bottomrule
\end{tabular}
}\end{center}
\end{table*}

\end{document}